\documentclass{article}

\usepackage[final,nonatbib]{neurips_2020}

\usepackage[utf8]{inputenc} 
\usepackage[T1]{fontenc}    
\usepackage{hyperref}       
\usepackage{url}            
\usepackage{booktabs}       
\usepackage{amsfonts}       
\usepackage{nicefrac}       
\usepackage{microtype}      
\usepackage{amsmath}
\usepackage{graphicx}
\usepackage{latexsym}  
\usepackage{verbatim}
\usepackage{multirow}
\usepackage{rotating}
\usepackage{wrapfig}
\usepackage{floatflt}
\usepackage{algorithm}
\usepackage{algorithmicx}
\usepackage[noend]{algpseudocode}
\usepackage{mathtools}
\usepackage{caption}
\usepackage[caption=false,subrefformat=parens,labelformat=parens]{subfig}
\usepackage{todonotes}
\usepackage{authblk}

\DeclareMathOperator{\KL}{D_{KL}}
\DeclareMathOperator{\Expect}{\mathbb{E}}

\title{Deep Transformers with Latent Depth}

\author[1]{Xian Li}
\author[2]{Asa Cooper Stickland}
\author[1]{Yuqing Tang}
\author[1]{Xiang Kong}

\affil[1]{Facebook AI \authorcr
  \{\tt xianl, yuqtang, xiangk\}@fb.com}
\affil[2]{University of Edinburgh \authorcr
  \{\tt a.cooper.stickland\}@ed.ac.uk}  
\begin{document}

\maketitle

\begin{abstract}

 
 The Transformer model has achieved state-of-the-art performance in many sequence modeling tasks. However, how to leverage model capacity with large or variable depths is still an open challenge. We present a probabilistic framework to automatically learn which layer(s) to use by learning the posterior distributions of layer selection. As an extension of this framework, we propose a novel method to train one shared Transformer network for multilingual machine translation with different layer selection posteriors for each language pair. The proposed method alleviates the vanishing gradient issue and enables stable training of deep Transformers (e.g. 100 layers). We evaluate on WMT English-German machine translation and masked language modeling tasks, where our method outperforms existing approaches for training deeper Transformers. Experiments on multilingual machine translation demonstrate that this approach can effectively leverage increased model capacity and bring universal improvement for both many-to-one and one-to-many translation with diverse language pairs.  

\end{abstract}

\section{Introduction}

The Transformer model has achieved the state-of-the-art performance on various natural language preprocessing (NLP) tasks, originally in neural machine translation \cite{vaswani2017attention}, and recently in massive multilingual machine translation \cite{arivazhagan2019massively,zhang2020improving}, crosslingual pretraining \cite{conneau2019unsupervised,liu2020multilingual}, and many other tasks. There has been a growing interest in increasing the model capacity of Transformers, which demonstrates improved performance on various sequence modeling and generation tasks \cite{yang2019xlnet, radford2019language, adiwardana2020towards}. 

Training Transformers with increased or variable depth is still an open problem. Depending on the position of layer norm sub-layer, backpropagating gradients through multiple layers may suffer from gradient vanishing \cite{ott2018scaling, wang2019learning, bachlechner2020rezero}. In addition, performance does not always improve by simply stacking up layers \cite{bapna2018training, wang2019learning}. When used for multilingual or multi-task pretraining, such as multilingual machine translation, crosslingual language modeling, etc., the simplicity of using one shared Transformer network for all languages (and tasks) is appealing. However, how to share model capacity among languages (and tasks) so as to facilitate positive transfer while mitigating negative transfer has not been well explored. 


In this work, we present a novel approach to train deep Transformers, in which the layers to be used (and shared) and the effective depth are not static, but learnt based on the underlying task. Concretely, we model the decision to use each layer as a latent variable, whose distribution is jointly learnt with the rest of the Transformer parameters. 
At training time we approximate the discrete choice with a Gumbel-Softmax \cite{jang2016categorical} distribution. The `soft weights' sampled from this distribution also act as gradient normalization for each layer, and this allows us to train very deep Transformers (up to 100 layers) without using regular layer normalization layers. At inference time, the learnt discrete choice can be used to directly derive a compact model by pruning layers with low probability, but we have the choice of leaving the learned layer selection probabilities as soft weights.
By evaluating on WMT'16 English-German machine translation (MT) and masked language modeling (MLM) tasks (similar to the XLM-R model \cite{conneau2019unsupervised}), we show that we can successfully train deeper Transformer (64-layer encoder/decoder model for MT, and 96-layer encoder for MLM) and outperform existing approaches in terms of quality and training stability.

We show this approach can be extended to learn task-specific sub-networks by learning different layer selection probabilities for each language pair in multilingual machine translation.
This result contributes to the growing interest of learning efficient architectures for multi-task and transfer learning in natural language understanding and generation \cite{pmlr-v97-stickland19a, pmlr-v97-houlsby19a, bapna-firat-2019-simple}. 


The main contributions of this paper are as follows. We present a probabilistic framework to learn which layers to select in the Transformer architecture. Based on this framework, we propose a novel method to train one shared Transformer network for multilingual machine translation with different layer selection probabilities for each language pair. The proposed method alleviates the vanishing gradient issue and enables stable training of deep Transformers. We conduct experiments on several tasks to evaluate the proposed approach: WMT'16 English-German machine translation, masked language modeling, and multilingual many-to-one as well as one-to-many machine translation with diverse languages. 

\section{Method}
\label{sec:method}

\paragraph{Background}


In this section, we briefly describe the standard Transformer layer architecture \cite{vaswani2017attention}.
For a hidden state $x_l$ of a single token at layer $l$, each Transformer layer is a function $F_l(x_l)$ that transforms its input $x_l$ by sequentially applying several sub-layers. The sub-layer is as follows:
\begin{equation}
     x_{l+1} = x_l + \mathrm{SubLayer}_{l}(\mathrm{Norm}(x_l)),
\end{equation}
where $\mathrm{SubLayer}_{l}(\cdot)$ is either a Self Attention module, an Encoder Attention module (for a Transformer decoder in a sequence-to-sequence model), or a feed-forward network (FFN) module, and $\mathrm{Norm}(\cdot)$ is a normalisation layer, usually layer-norm \cite{ba2016layer}. This is the `pre-norm' setting which is now widely used \cite{ott2018scaling}, as opposed to `post-norm' in which case $\mathrm{Norm}(\cdot)$ would be applied after the residual connection: $x_{l+1} = \mathrm{Norm}(x_l + \mathrm{SubLayer}_{l}(x_l))$.


\begin{figure}[t]
\vskip 0.2in
\begin{center}
\centerline{\includegraphics[width=\linewidth]{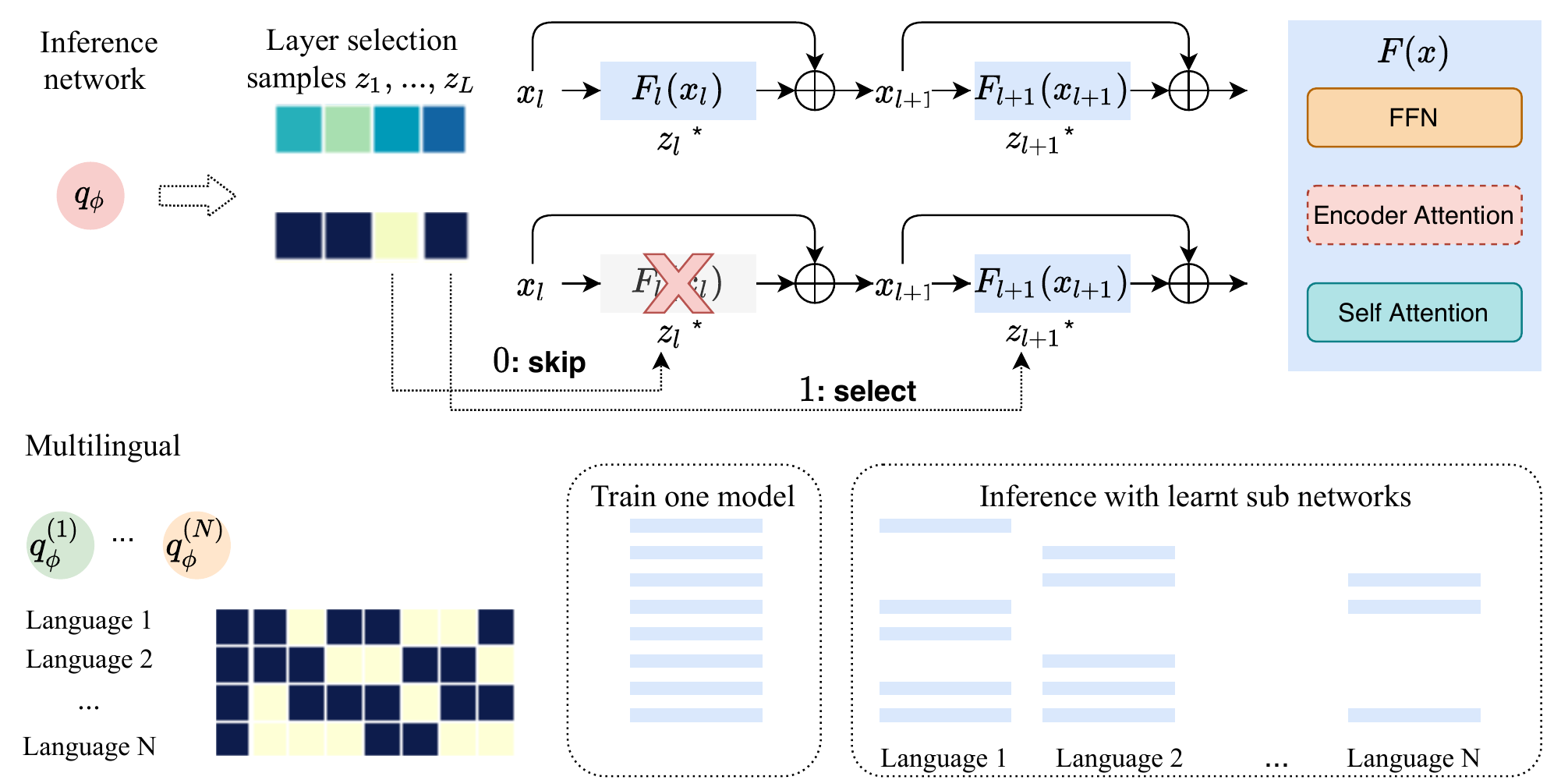}}
\caption{We learn the posterior distribution $q_{\phi}$ to ``select" or ``skip" each layer in Transformers. In multilingual setting, each language learns their own ``views" of latent layers in a shared Transformer. }
\label{fig:layer_select}
\end{center}
\vskip -0.2in
\end{figure}

\subsection{Latent Layer Selection}
For each Transformer layer $l$, we treat the selection of all sub-layers in non-residual block $F_l(x)$ as a latent variable $z_l$ from a parameterizable distribution $p(z)$,
\begin{equation}
     x_{l+1} = x_l + z_l \times F_{l}(x_l) \text{,   } z_l \sim p(z;l) 
\label{eq:ll}
\end{equation}

where the standard Transformer \cite{vaswani2017attention} is a special case with $z_l=1$ for $l=0, ..., L-1$, where $L$ is the depth of the network, i.e. total number of layers.

For the sequence generation task $p(y\mid x)$ parameterized by a Transformer network with the remaining standard parameters ${\Theta}$, we assume the following generative process:
\begin{equation}
\label{eq:loglik}
     y \sim p(y\mid x; \theta, z) ,~~
     p(y\mid x) = \int_{z}p(y\mid x;\Theta, z) p(\Theta, z)\, \mathrm{d}\Theta\mathrm{d}z 
\end{equation}


\paragraph{Parameterization and inference of $z$.} We model $z_l$ as discrete latent variable from a Bernoulli distribution with  $ z_l \sim \mathcal{B}(\pi;l) \text{, } \pi \in [0, 1]$
indicating select or skip the non-residual block $F_l(x)$ in layer $l$, and samples from one layer are independent from other layers. This modeling choice allows us to prune layers which reduces inference cost and may regularize training. 

Marginalizing over $z$ becomes intractable when $l$ grows large. Therefore, we use variational inference as a more general optimization solution. Specifically, we 
instead maximize the evidence lower bound (ELBO) of Eq. \ref{eq:loglik} 
\begin{equation}
   \log p(y \mid x) \geq \Expect_{q_{\phi}(z)} [\log p_\theta(y\mid x, z)] 
    - \KL(q_{\phi}(z) \parallel p(z)) 
\end{equation}

We point out that although we could treat the rest of the network parameters $\Theta$ as latent variables too and model the joint distribution of $p(\theta, z)$, which could be optimized using Coupled Variational Bayes (CVB) and optimization embedding as demonstrated in \cite{shaw2019meta} for neural architecture search, in practice we found a simpler optimization procedure (Algorithm \ref{alg:main}) to learn both $\theta$ and $z$ jointly from scratch. 

We use the Gumbel-Softmax reparameterization \cite{jang2016categorical} to sample from the approximate posterior $q_{\phi}(z)$ which makes the model end-to-end differentiable while learning (approximately) discrete policies without resorting to policy gradients. To allow both ``soft weighting" and ``hard selection" of layers, each of which has the appealing property of achieving model pruning while training with larger model capacity, we generate soft samples of $z$ during training and draw hard samples for pruning at inference time if $q_{\phi}(z)$ becomes (close to) discrete. We directly learn the logits parameter $\alpha_l$ for each layer $l$:
\begin{equation}
\label{eq:z_sample}
    z_l^i(\alpha_l) = \frac{\exp((\alpha_l(i) + \epsilon(i))/\tau)}{\sum_{i\in\{0,1\}} \exp((\alpha_l(i) + \epsilon(i))/\tau)} \text{ , } \epsilon \sim \mathcal{G}(0, 1)
\end{equation}
where $\mathcal{G}(0, 1)$ is the Gumbel distribution, and $\tau$ is a temperature hyperparameter which increases the  discreteness of samples when $\tau \to 0$. 
For $p(z)$ we can use the conjugate prior Beta$(a, b)$ which allows us to express different preferences of $z$, such as $a=b=1$ for an uniform prior, $a>b$ to bias towards layer selection and $a<b$ to favor skipping layers.

\paragraph{Gradient scaling.} Next we analyze the impact of latent layers on gradient backpropagation during training in the pre-norm setting. In Eq. \ref{eq:grad}, we can see that given the forward pass loss $\mathcal{L}$, the gradient accumulation from higher layers $m_{l<m<L}$ is now weighted by the their corresponding latent samples $z_m$, which acts as gradient scaling. In Section \ref{sec:exp} we show that with such gradient normalization we can train deeper Transformers without using layer normalisation.  

\begin{equation}
\label{eq:grad}
    \frac{\partial \mathcal{L}}{\partial x_l} = \frac{\partial \mathcal{L}}{\partial x_L} \times (1+\sum_{m=l}^{L-1}z_m \frac{\partial F_m(x_m)}{\partial x_l})
\end{equation}

\subsection{Multilingual Latent Layers}
\label{sec:multilatent}
It is sometimes convenient to share a Transformer network across multiple languages, enabling crosslingual transfer, with recent success in multilingual machine translation and multilingual pretraining (e.g.\ multilingual BERT and BART) \cite{arivazhagan2019massively,conneau2019unsupervised,pires2019multilingual,liu2020multilingual}. 
Current approaches share a vanilla (usually 12-layer) Transformer across all languages.

To explore the potential of latent layers for a multilingual Transformer, we let each language learn its own layer utilization given a single Transformer network $\Theta$ shared among $N$ languages by learning its own posterior inference network $q_{\phi}^{(n)}$ of $\{\alpha_l\}$. We acknowledge that an alternative is to learn a shared inference network $q_{\phi}(n)$ which takes language $n$ as input. The latter may enable learning commonalities across languages but at the cost of extra parameters, including a non-trivial $N \times d$ parameters for language embeddings. Therefore, we chose the former approach and leave the latter (and the comparison) for future work. With this modeling choice, we can still encourage layer-sharing across languages by using the aggregated posterior across languages $\tilde{q}(z)$ as the prior in the $D_{KL}$ term: 
\begin{equation}
\label{eq:agg_kl}
    \KL (q_{\phi}(z)\parallel \tilde{q}(z)) = \Expect_{q_{\phi}(z)}[ \log \frac{q_{\phi}(z)}{\tilde{q}(z)} ]\text{ , } \tilde{q}(z)=\frac{1}{N}  \sum_{n=1}^N q_{\phi}(z\mid x^{(n)}, y^{(n)}, \hat{\theta})
\end{equation}
\paragraph{Latent Layers with Targeted Depth}
To deploy Transformers in the real world, we would like to have lower computational cost at inference time. Within a Transformer layer, some computation is parallel, such as multi-head attention, but the time and space complexity at inference time grows linearly with the number of layers. Therefore, pruning layers at test time directly reduces inference cost.
Our approach can be extended to perform model pruning, encouraging the model to achieve a target depth $K$ by adding an extra loss $\mathcal{L}_{K} = \|\sum_{l=0}^{L-1} u_l  - K \|_2$
where $u_l$ refers to the ``utilization" of layer $l$. $u_l$ can be approximated by samples of the latent variables $z_l$ and for the multilingual case $u_l=\sum_{n=1}^N z_l^{(n)}/N$.

The general loss for training a Transformer with latent depth $K$ is
\begin{equation}
\label{eq:ll_loss}
 \mathcal{L}_{LL} = \underbrace{ \Expect_{q_{\phi}(z)} [-\log p_\theta(y\mid x, z)] + \beta \KL(q_{\phi}(z) \parallel p(z))  }_{\mathcal{L}_{\text{ELBO}}} + \lambda  \mathcal{L}_{K}
\end{equation}

To learn $\Theta$ and $q_{\phi}$ jointly from scratch, we use an two-level optimization procedure described in Algorithm \ref{alg:main}. This training strategy is inspired by the Generalized Inner Loop Meta-learning \cite{grefenstette2019generalized}. We provide a more detailed explanation of this training procedure in Appendix \ref{app:training}.

\begin{wrapfigure}{R}{0.5\textwidth}
\vspace{-5mm}
    \begin{minipage}{0.5\textwidth}
\begin{algorithm}[H]
\caption{\label{alg:main} Training with Latent Layers}
\begin{algorithmic}[1]
    \State Initialize  $\Theta$, $q_{\phi}$. 
    \For{t=1, ..., T}
        \For{i=1, ..., I}
                \State Sample a mini-batch $(x, y) \sim D$ .
                \State{Sample $z_{l=0, ..., L-1}$ with Eq. \ref{eq:z_sample} 
                }
                \State{Compute  $\hat{\mathcal{L}}_{LL}((x, y);\Theta_{i-1}, q_{\phi}^{t-1})$ with Eq. \ref{eq:ll_loss}}
                \State{Update $\Theta_i=\Theta_{i-1} - \eta \nabla_{\Theta_{i-1}} \hat{\mathcal{L}}_{LL}$ }
        \EndFor
		\State{Update $q_{\phi}^t=q_{\phi}^{t-1} - \eta \nabla_{q_{\phi}^{t-1}} \hat{\mathcal{L}}_{LL} $}
	\EndFor
\end{algorithmic}
\end{algorithm}
  \end{minipage}
  \vspace{-8mm}
  \end{wrapfigure}

\section{Experimental Settings}
\label{sec:exp}

We first evaluate on the standard WMT English-German translation task and a masked language modeling task to demonstrate the effectiveness of the proposed approach at enabling training deeper Transformers and whether this increased depth improves model performance. We then evaluate multilingual latent layers (see section~\ref{sec:multilatent}) on multilingual machine translation.


\paragraph{Bilingual Machine Translation.} We use the same preprocessed WMT'16 English-German sentence pairs as is used in \cite{vaswani2017attention, wang2019learning}. To make comparison more clear and fair, we evaluate on the last model checkpoint instead of ensembles from averaging the last 5 checkpoints. We use beam size 5 and length penalty 1.0 in decoding and report corpus-level BLEU with sacreBLEU \cite{post2018call}.

\paragraph{Crosslingual Masked Language Modelling.} 

We test our method on a scaled-down version of XLM-R \cite{conneau2019unsupervised}, intending to show the promise of our method, but not obtain state-of-the-art results on downstream tasks.
In particular we use as training data the Wikipedia text of the 25 languages used in the mBART \cite{liu2020multilingual} model, and evaluate using perplexity on a held out dataset consisting of 5000 sentences in each language (sampled randomly from each Wikipedia text).

\paragraph{Multilingual Machine Translation.}  We evaluate the proposed approach on multilingual machine translation using the 58-language TED corpus~\cite{qi2018and}. To study its performance independent of task similarity and difficulty, we evaluate on both \textit{related} (four low resource languages and four high resource languages from the same language family) and \textit{diverse} (four low resource languages and four high resource ones without shared linguistic properties) settings as is used in \cite{wang2020balancing}.
Dataset descriptions and statistics are summarized in the Appendix \ref{app:data}. For each set of languages, we evaluate both many-to-one (M2O), i.e. translating all languages to English, and one-to-many (O2M), translating English to each of the target languages, which is a more difficult task given the diversity of target-side languages. 

\paragraph{Baselines.} We compare to the standard Transformer with static depth on machine translation task and ``wide" model, e.g.  Transformer-big architecture in \cite{vaswani2017attention} which increases the hidden (and FFN) dimension and has been a common approach to leverage large model capacity without encountering the optimization challenges of training a deeper model. 

We also compare to recent approaches to training deeper Transformers:
\begin{itemize}
     \item Random Layer drop. For deeper models where the static depth baselines diverged, we apply the random LayerDrop described in \cite{fan2019reducing} which trains a shallower model by skipping layers.
    \item Dynamic linear combination of layers (DLCL). 
    This is a recently proposed approach to address vanishing gradient by applying dense connections between layer which was demonstrated effective for machine translation\cite{wang2019learning}. 
    \item ReZero\cite{bachlechner2020rezero}. This is similar to our method in that both methods learn to weigh each layer. 
    The key difference is that ReZero learns (unconstrained) weighting parameters. In our experiments, we found ReZero suffers from gradient exploding and training loss diverged. 
\end{itemize}




\section{Results}
\begin{figure}[!t]
\vspace{-3mm}
\centering
\subfloat[Gradient norms of encoder and decoder in standard Transformer.]{
\includegraphics[width=1.0\linewidth]{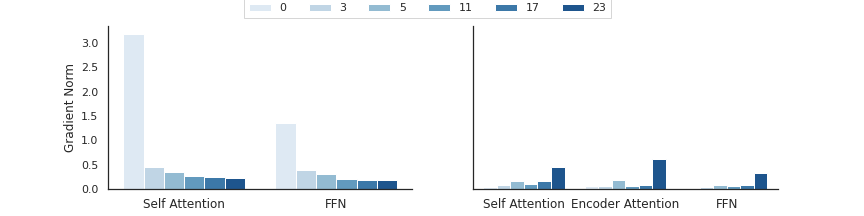}
\label{fig:gnorm_base}
}
\smallskip\par
\vspace{-3mm}
\subfloat[Improvement of decoder's gradient norm using latent layers.]{
\includegraphics[width=0.95\linewidth]{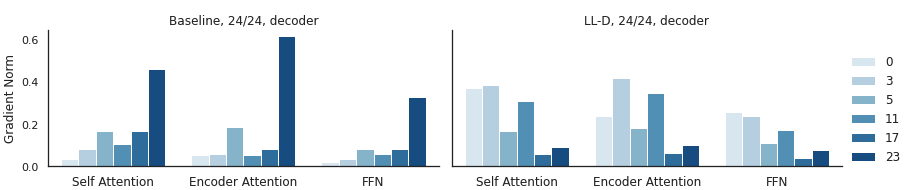}
\label{fig:gnorm_dec}
}
\vspace{-2mm}
\caption{Comparing gradient norms of baseline (a) and using latent layers (b).}
\vspace{-2mm}
\label{fig:gnorm_both}
\end{figure}

\begin{wrapfigure}{rt}{0.5\textwidth}
\vspace{-23mm}
\centering
\includegraphics[width=\linewidth]{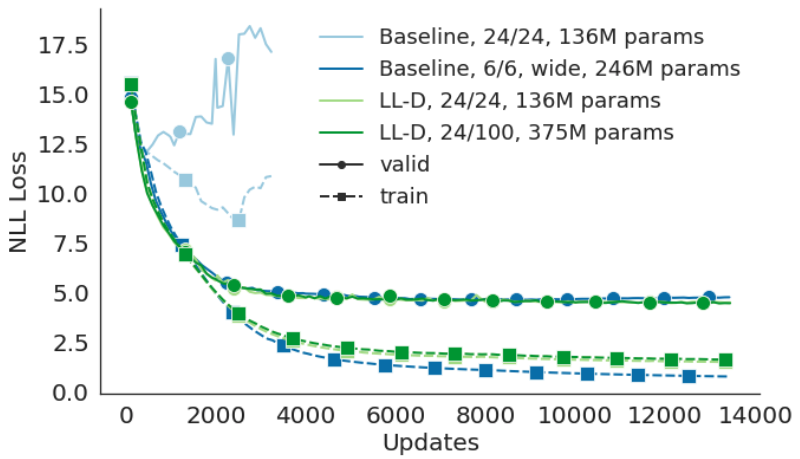}
\vspace{-6mm}
\caption{\small Comparing learning curves, training and validation per-token negative loglikelihood (NLL) loss, of baseline models (static depth) and the proposed method when training deeper model (decoder).}
\vspace{-6mm}
\label{fig:nll_loss}
\end{wrapfigure}
\subsection{Addressing vanishing gradient}
First, we empirically show that with static depth, gradient vanishing happens at bottom layers of decoder Figure \ref{fig:gnorm_base}. The effect of training with latent layers using the proposed approach is illustrated in Figure \ref{fig:gnorm_dec}, which shows that gradient norms for bottom layers in the decoder are increased.


Next, we compared the learning curves when training deeper models.  As is shown in Figure \ref{fig:nll_loss} (evaluated on multilingual translation task O2M-Diverse dataset), the baseline model with static depth diverged for a 24-layer decoder, while using the latent layers ((LL-D) approach we could train both 24-layer and 100-layer decoder successfully. We further compared the 100-layer model with a wider model (Transformer-big), and found that besides stable training, deep latent layer models are less prone to overfitting (i.e.\ they achieve lower validation loss, with a smaller gap between train and validation losses) despite having more parameters.  


\subsection{En-De Machine Translation}
In Table \ref{tab:wmt} we evaluate on training deeper Transformers and examine the impact of   
\begin{wraptable}{r}{0.7\textwidth}
\small
\vspace{-2mm}
    \centering
    \begin{tabular}{lccccc}
    \toprule
    Model & Params &  $\text{NLL}_{\text{valid}}\downarrow$  &  $\text{BLEU}_{valid}\uparrow$  &  $\text{BLEU}_{test}\uparrow$ \\
    \hline
    Transformer-Big  & 246M &	2.081 &28.7 &28.1   \\
    DLCL, 36/36 & 224M & 2.128 &28.5 & 27.7 \\
    DLCL, 48/48 & 224M & 2.090 & \textbf{28.8} & 28.1 \\
    \hline
    LL-D, 12/24 & 135M  &	2.179 &28.1$\pm$0.08 & 27.2$\pm$0.04	 \\
    LL-D,12/48 & 211M  &	2.128 & 28.1$\pm$0.00  & 27.3$\pm$0.04	 \\
    LL-Both, 36/36 & 224M  &	2.147 & 28.4$\pm$0.07  & 28.1$\pm$0.07	 \\
    LL-Both, 48/48 & 287M  &	2.078 & 28.7$\pm$0.10   & \textbf{28.7$\pm$0.09}	 \\
     LL-Both, 64/64 & 371M  &	\textbf{2.069} & 28.5$\pm$0.07 & 28.4$\pm$0.08	 \\
    \bottomrule
    \end{tabular}
    \caption{\small Performance on WMT'16 En-De. For BLEU scores evaluation, we provide standard errors from 5 runs with different seeds.}\label{tab:wmt}
\vspace{-9mm}
\end{wraptable}
latent layers in decoder (LL-D) and both encoder and decoder (LL-Both) respectively. Compared to existing methods for training deeper Transformers such as using dense residual connections (DLCL), our approach can leverage larger model capacity from increased depth and achieved improved generalization.

\subsection{Masked Language Modeling}
\begin{wraptable}{r}{0.4\textwidth}
\small
\vspace{-5mm}
    \centering
    \begin{tabular}{lcc}
    \toprule
    Model & Params &  Perplexity $\downarrow$  \\
    \hline
    Static depth 24  & 202M & 2.91	  \\
    LL, 24  & 202M & 2.82	  \\
    Static 48 & 372M & 2.60  \\
    LL, 48  & 372M & 2.71  \\
    Static 96 & 712M & Diverged \\
     + layer-drop & 712M & Diverged \\
    LL, 96  & 712M &  2.66 \\

    \bottomrule
    \end{tabular}
    \caption{Perplexity on held-out data for crosslingual masked language modeling.}\label{tab:xlm}
    \vspace{-8mm}
\end{wraptable}
Latent layers (LL) is also shown to be effective for training deeper encoder without divergence (see Table~\ref{tab:xlm}). For 24 and 48 layer encoders, we observed stable training with 2x learning rate and achieved better performance for 24 layers. However the result of scaling up to 96 layers was slightly worse performance than a vanilla 48 layer model. This shows the promise of the method for stabilising training at increased depth, however we did not attempt to scale up our data to match our larger model capacity. 

\subsection{Multilingual Translation}
We separately test the impact of applying latent layers in the decoder (LL-D), encoder (LL-E) and both (LL-Both).

\begin{table*}[h]
\centering
\begin{tabular}{lcccccccccccccr}
\toprule
  Model &  Params & Avg.  &  aze & bel &  ces & glg &  por &  rus &  slk & tur    \\
\hline
 6/6 & 63.6M & 19.65 &  5.4 &  9.1 &  21.9 & 22.4 &  38.6 & 19.4 & 24.6 &  15.8  \\
6/6, wide & 190M & 20.33 & {\bf 5.7} & 9.7 & 22.4 & 23.1 & 40.3 & 20.6 & 24.1 & 16.8  \\
 12/12 & 95.1M  &  20.48 &  5.6 &  10.3 &  23.1 & 22.8 &  39.7 & 20.1 & {\bf 25.1} &  17.1  \\
 12/24  & 133M & NA & - & - & - & - & - & -& - & -  \\
 24/24 & 158M & NA & - & - & - & - & - & -& - & -  \\
 +layer drop &  158M & 11.16 & 3.3 & 7.5 & 11.6 & 14.4 & 23.4 & 10.4 & 12.9 & 5.8  \\
\hline
 LL-D, 12/24 & 133M & 20.83 & 5 & 10.2 & 23.4 & {\bf 24.3} & 40.3 & {\bf 21}  & 24.8 &  {\bf 17.6} \\
 LL-D,  24/24 & 158M & {\bf 20.84} & 5.3 & {\bf 10.6} & {\bf23.4} & 23.7 & {\bf 40.7} & 20.9  & 24.8 &  17.5 \\
\bottomrule
\end{tabular}
\caption{BLEU scores for one-to-many multilingual translation on related languages. ``NA" means training diverged.}
\label{tab:t8r-o2m}
\vspace{-3mm}
\end{table*}



\paragraph{Latent layers in decoder.} To evaluate the impact of increased decoder depth, we tested on one-to-many (O2M) multilingual translation. In Table \ref{tab:t8r-o2m} we show performance on the ``Related" languages setting. Baseline models began to diverge when decoder depth increases to $L=24$, and applying random LayerDrop did not help. 
Latent layers allows us to train the same depth successfully, and we observe improvement in translation quality for both language pairs as well as overall quality shown by the average BLEU score.
In Table \ref{tab:t8d-o2m}, we evaluate the impact of deeper decoder with latent layers in the O2M-Diverse setting. This is a more challenging task than O2M-Related since decoder needs to handle more diversified syntax and input tokens. 

\begin{table*}[h]
\centering
\begin{tabular}{lccccccccccccr}
\toprule
 Model & Avg.  &  bos & mar &  hin & mkd &  ell & bul &  fra &  kor    \\
\hline
6/6  &  22.12 &  12.6 &  11.1 &  14.6 & 22.7 &  29.8 & 31.8 & 37.3 &  17.1  \\
 6/6, wide & 23.51 & 12.7 &11.3 & 13.9& 23.8 & 32.5 & 34.8 & 40.6 & 18.5 \\
 12/12   &  23.34 &  13.1 &  11.1 &  13.6 & 22.5 &  32.7 & 34.7 & 40.4 &  18.6  \\
 12/24  & NA & - & - & - & - & - & -& - & -  \\
 24/24  & NA & - & - & - & - & - & -& - & -  \\
 +layer drop & 22.06 & 13.0 & 10.0 & 12.2 & 21.5 & 30.7 & 33.0& 38.5 & 17.6  \\
\hline
LL-D, 12/24 & 23.70 & 13.4 & 10.7 & 14.1 & 22.8 & 33.1 & 35.1  & 41.1 &  19.3 \\
LL-D,  12/100 & 24.16 & 13.5 & 10.6 & 13.8 &  24.1 & 32.7  & {\bf 38.2} & 41.3 & 19.1  \\
LL-D,  24/24  &{\bf 24.46}& {\bf 15.5} & {\bf 11.4} & {\bf 14.6} & {\bf 24.4} & {\bf 33.5}&  35.5 & {\bf 41.5}& {\bf 19.3} \\

\bottomrule
\end{tabular}

\caption{\small BLEU scores for one-to-many multilingual translation on diverse languages.}
\label{tab:t8d-o2m}
\vspace{-5mm}
\end{table*}
\paragraph{Latent layers in encoder, decoder, and both.} We use the many-to-one multilingual translation task to verify the pattern observed above, and test the effect of increased depth in encoder. Results are summarized in Table \ref{tab:t8d-m2o}. Similar to O2M, standard Transformer begins to diverge when decoder depth increase over 24 while applying latent layers enable successful training and yields improved translation quality. 

\begin{table*}[h]
\centering
\begin{tabular}{lccccccccccccr}
\toprule
 Model & Avg.  &  bos & mar &  hin & mkd &  ell & bul &  fra &  kor    \\
\hline
6/6  &  25.95 &  20.7 &  8.6 &  19.2 & 30.0 &  36.3 & 36.9& 38.4 &  17.5  \\
12/12  &  27.73 &  22.5 &  9.4 &  20.1 & 31.6 &  38 & 39.6 & 40.8 &  19.9  \\
24/12  & 27.86 & 23.7  & 9.7 &  21.6  &  31.2  & 37.6   & 39.3  &  40.0  &  19.8  \\
24/24  & NA & - & - & - & - & - & -& - & -  \\
+layer drop & 26.7 & 21.3 & 9 & 19.2 & 29.2 & 37.5 & 38.8& 39.9 & 18.7  \\
\hline
LL-E, 36/12   & 27.98 & {\bf 24.2} & 10.2 &  21.9 &  32 &  37.3& 38.8 & 39.3 &  20.1 \\
LL-D, 12/24  & 27.63 & 22.4 & 9.3 & 20.2 & 30.8 & 38.2 & 39.7 & 40.5 & 19.9  \\
LL-D, 12/36  & 27.89 & 22.3 & 9.5 & 21.1 & 30.7 & 38.2 & 40.2 & {\bf 41.2} & 19.9  \\
LL-D, 24/24   & 28.43&  23.6 & 10.0 &  21.9 &  31.7 & {\bf 38.4}& {\bf 40.3} & {\bf 41.2}& 20.4 \\
LL-Both, 24/24  & {\bf 28.56} & 23.5 & 10.3 & {\bf 22.3} & {\bf 32.8} & 38.3 & 40 & 40.8 & {\bf 20.5}  \\
\bottomrule
\end{tabular}
\caption{\small BLEU scores of models with increased depth in the encoder and decoder for many-to-one on diverse languages.}
\label{tab:t8d-m2o}
\vspace{-3mm}
\end{table*}

\begin{wrapfigure}{r}{0.7\textwidth}
\vspace{-7mm}
\centering
\includegraphics[width=\linewidth]{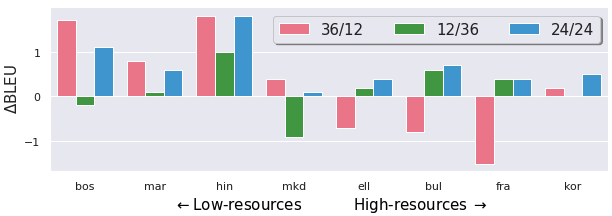}
\caption{Quality improvement (over static depths 12/12) by allocating increased capacity to all-encoder (36/12), all-decoder (12/36), and even allocation (24/24).}
\label{fig:deep_end}
\vspace{-8mm}
\end{wrapfigure}
By applying latent layers to encoder only (LL-E) we found increased depth (36/12) improves low resource languages (e.g. bos and hin) over the baseline (12/12). However, deeper decoder (12/36) or even allocation of depth (24/24) brings consistent gains as is shown in Fig \ref{fig:deep_end}.
\section{Analysis}
\label{sec:analysis}

In this section, we analyze the effect of several modeling choices and understand their contribution to the results.

\begin{figure}[ht]
\vspace{-4mm}
\begin{center}
\centerline{\includegraphics[width=\linewidth]{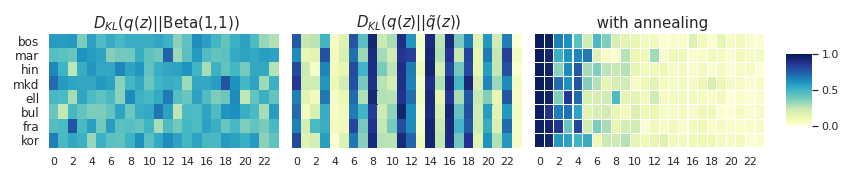}}
\caption{Layer selection samples $z_l$ at epoch 1 from different priors used for $D_{KL}$.}
\label{fig:priors}
\end{center}
\vspace{-5mm}
\end{figure}

\paragraph{Effect of Priors}
In Figure \ref{fig:priors} we illustrate the difference between using aggregated posterior $\tilde{q}(z)$ versus a uniform prior $\text{Beta}(1,1)$ in computing the $\KL$ loss. 

Compared to the uniform prior, using the aggregated posterior as prior discretizes layer utilization, that is, the model is incentivised to make layer selections consistent across all languages, i.e. facilitating parameter sharing. Interestingly, the learnt ``sharing" pattern by using $\tilde{q}(z)$ as prior is consistent with heuristics such as dropping every other layer for pruning which was empirically found effective \cite{fan2019reducing}. However, training with such a prior in the beginning can lead to ``posterior collapse'', which is a well-known challenge found in training variational autoencoders. After applying ``KL annealing'' (annealing the $\KL$ coefficient $\beta$), we can see that layer selection samples are more continuous with a curriculum to use the bottom layers first. 

\begin{figure}[t]
\vskip 0.2in
\begin{center}
\centerline{\includegraphics[width=\linewidth]{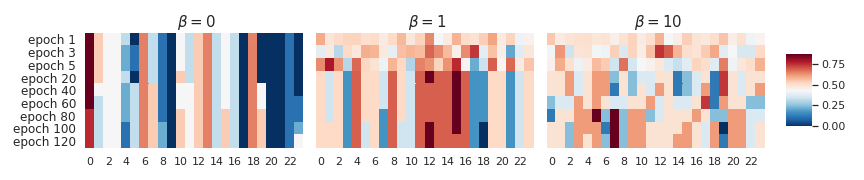}}
\caption{Visualization of layer utilization $u_l$ during training using the M2O Diverse dataset.}
\label{fig:layer_util}
\end{center}
\vspace{-3mm}
\end{figure}

\begin{wraptable}{r}{0.4\textwidth}
\small
\vspace{-4mm}
    \centering
    \begin{tabular}{lcc}
    \toprule
     & $\Expect_{L}$ & Avg. valid BLEU \\
    \hline
    $\beta=0$  & 10.25 &	28.50   \\
   $\beta=1$  & 11.25 &	28.53   \\
   $\beta=10$  & 12.125 &	28.23   \\
    \bottomrule
    \end{tabular}
    \caption{\small Impact of the KL coefficient $\beta$ on network effective depth ($\Expect_{L}$) and translation quality, evaluated on M2O-Diverse.}\label{tab:beta}
\vspace{-3mm}
\end{wraptable}
\paragraph{Effect of $\beta$.} In order to understand how the $\KL$ loss term affects layer selection policies and samples \textit{throughout} training, we vary the $\KL$ coefficient $\beta \in \{0,1,10\}$. First, we examine layer utilization $u_l$, e.g. whether ``hot layers" ($u_l \to 1$) and ``cold layers" ($u_l \to 0$) change over time. As is shown in Figure \ref{fig:layer_util}, without the $\KL$ term, layer utilization stays constant for most of the layers, especially several top layers whose parameters were rarely updated. By increasing the contribution from the $\KL$ to the total loss, layer selections are more evenly spread out across languages, i.e. $u_l$ becomes more uniform. This is also reflected in Table \ref{tab:beta} where the ``effective depth" $\Expect_{L}$ increases with $\beta$.

\subsection{Ablation Studies}
In this section, we provide ablation experiments to understand how different loss terms contribute to the results.  Table \ref{tab:abl-o2m} compares the effect on translation quality from different loss terms in Eq. \ref{eq:ll_loss}. We can see that optimizing the $\mathcal{L}_{ELBO}$ loss brings the most quality gains, and $\mathcal{L}_K$ loss adds additional improvement by acting as a regularization. 

\begin{table*}[h]
\centering
\begin{tabular}{lccccccccccccr}
\toprule
 Model & Avg.  &  bos & mar &  hin & mkd &  ell & bul &  fra &  kor    \\
\hline
LL-D, 24/24  &{\bf 24.46}& {\bf 15.5} & {\bf 11.4} &  14.6 &  24.4 & {\bf 33.5}& {\bf 35.5} &  41.5& {\bf 19.3} \\ 
- $\mathcal{L}_K$  & 24.28 & 14.5 & 10.8 & 14.3 & {\bf 25} & 33.4 & 35.4 & {\bf 41.6} & 19.2 \\
- $\KL$  & 23.89 & 13.7 & 11.0 & {\bf 14.7} & 24.2 & 32.9 & 35.0 & 40.9 & 18.9 \\
- both  & 23.75 & 13.4 &  10.9 & 14.2 & 23.6 & 33.2 & 35.2 & 40.7 & 18.8  \\
\bottomrule
\end{tabular}
\caption{\small Effects from different terms in $\mathcal{L}_{LL}$ evaluated on the O2M-Diverse dataset.}
\label{tab:abl-o2m}
\end{table*}
\subsection{Latent depth vs. static depth}
We compare a deeper model with latent effective depth $\Expect[L]$ to models with the same depth trained from scratch. 

\begin{wraptable}{r}{0.65\textwidth}
\small
    \centering
    \begin{tabular}{lccc}
    \toprule
    Model &   $\text{BLEU}_{valid}\uparrow$  &  $\text{BLEU}_{test}\uparrow$ \\
    \hline
    Latent depth, $L=24$, $\Expect[L]=12$ & \bf{28.6$\pm$0.07}  & \bf{27.88$\pm$0.04}	 \\
    Static depth, $L=12$  &27.2 & 26.5 \\
    \bottomrule
    \end{tabular}
    \caption{\small Comparing a 24 latent layers model with effective depth $\Expect[L]=12$ with a 12-layer static depth model trained from scratch, evaluated on WMT'16 En-De.}\label{apptab:wmt}
\vspace{-5mm}
\end{wraptable}
As is observed in both bilingual  (Table \ref{apptab:wmt}) and multilingual (Table \ref{tab:abl-m2o}) machine translation tasks, training a deeper model with latent depth outperforms standard Transformer with the same number of effective layers but trained  with static depth. 

\begin{table*}[h]
\centering
\begin{tabular}{lccccccccccccr}
\toprule
 Model & Avg.  &  bos & mar &  hin & mkd &  ell & bul &  fra &  kor    \\
\hline
Latent depth, $\Expect[L]=14.5$  &{\bf 28.43}&  23.6 & 10.0 & {\bf 21.9} & {\bf 31.7} & {\bf 38.4}& {\bf 40.3} & {\bf 41.2}& {\bf 20.4} \\
 Static depth, $L=15$  & 27.9 & {\bf 23.9} & {\bf 10.3} & 21.5 & 31.4 & 37.5 & 38.9& 39.8 & 19.9 \\
\bottomrule
\end{tabular}
\caption{Comparing a 24 latent layers model with effective depth $\Expect[L]=14.5$ with a 15-layer static depth model trained from scratch, evaluated on M2O-Diverse dataset.}
\label{tab:abl-m2o}
\vspace{-3mm}
\end{table*}

\section{Related Work}
The Transformer model~\cite{vaswani2017attention} has achieved state-of-the-art performance on various natural language processing (NLP) tasks. Theoretical results suggest networks often have an expressive power that scales exponentially in depth instead of width~\cite{poole2016exponential}, and recent work~\cite{zhang2019improving,adiwardana2020towards,qi2018and,wang2020balancing} finds that deeper Transformers improve performance on various generation tasks. 
However, deeper Transformer models also face the gradient vanishing/exploding problem leading to unstable training~\cite{bapna2018training,wang2019learning}. In order to mitigate this issue, Huang et al. (2016)~\cite{huang2016deep} drop a subset of layers during the training, and bypass them with the identity function. 
Zhang et al. (2019)~\cite{zhang2019fixup} propose an initialization method to scale gradients at the beginning of training to prevent exploding or vanishing gradient. Bachlechner et al. (2020)~\cite{bachlechner2020rezero} initialize an arbitrary layer as the identity map, using a single additional learned parameter per layer to dynamically facilitates well-behaved gradients and arbitrarily deep signal propagation. Fan et al. (2019)~\cite{fan2019reducing} introduce a form of structured dropout, \emph{LayerDrop}, which has a regularization effect during training and allows for efficient pruning at inference time. Concurrent work which shown improvement on NMT task by increasing model depth includes Zhang et al. (2020) ~\cite{zhang2020improving} and Wei et al. (2020) ~\cite{wei2020multiscale}.

Exploring dynamic model architecture beyond hard parameter sharing has received growing interest. In multi-task learning, Multi-Task Attention Network (MTAN) ~\cite{liu2019end}, routing network ~\cite{rosenbaum2017routing} and branched network ~\cite{vandenhende2019branched} enables soft parameter sharing by learning a dynamic sub-network for a given task. One concurrent work ``GShard" ~\cite{lepikhin2020gshard} also demonstrate deeper model with conditional computation brings consistent quality improvement for multilingual translation. More work on learning an adaptive sub-network includes BlockDrop~\cite{wu2018blockdrop} which learns dynamic inference paths per instance, and SpotTune ~\cite{guo2019spottune} which learns which layers to finetune or freeze to improve transfer learning from a pretrained model.





\section{Conclusion}
We proposed a novel method to enable training deep Transformers, which learns the effective network depth, by modelling the choice to use each layer as a latent variable.
Experiments on machine translation and masked language modeling demonstrate that this approach is effective in leveraging increased model capacity and achieves improved quality. We also presented a variant of this method in a multilingual setting where each language can learn its own sub-network with controllable parameter sharing. This approach can be extended to use a shared Transformer for multi-task learning in NLP tasks, and offers insight into which layers are important for which tasks.

\section*{Broader Impact}
This work proposes a new method to leverage a model with increased depth during training, while learning a compact sub-work with reduced depth which can be used for deployment in real-world applications where Transformers have achieved state-of-the-art quality such as machine translation systems, dialog and assistant applications, etc, as reducing the number of layers especially in decoder (often autoregressive) can have direct impact on reducing inference-time latency, memory consumption, etc. However scaling up the number of layers adds to energy cost of training, even if we can prune at inference time.

We hope our research on multilingual NLP will contribute to the effort of improving the standard of NLP tools for low-resource languages. However we only test our machine translation systems on to-English or from-English tasks, leaving out translation from non-English languages to other non-English languages entirely. 

\medskip

\small
\bibliographystyle{plain}
\bibliography{neurips_2020}

\begin{thebibliography}{10}

\bibitem{adiwardana2020towards}
Daniel Adiwardana, Minh-Thang Luong, David~R So, Jamie Hall, Noah Fiedel, Romal
  Thoppilan, Zi~Yang, Apoorv Kulshreshtha, Gaurav Nemade, Yifeng Lu, et~al.
\newblock Towards a human-like open-domain chatbot.
\newblock {\em arXiv preprint arXiv:2001.09977}, 2020.

\bibitem{antoniou2018train}
Antreas Antoniou, Harrison Edwards, and Amos Storkey.
\newblock How to train your maml.
\newblock {\em arXiv preprint arXiv:1810.09502}, 2018.

\bibitem{arivazhagan2019massively}
Naveen Arivazhagan, Ankur Bapna, Orhan Firat, Dmitry Lepikhin, Melvin Johnson,
  Maxim Krikun, Mia~Xu Chen, Yuan Cao, George Foster, Colin Cherry, et~al.
\newblock Massively multilingual neural machine translation in the wild:
  Findings and challenges.
\newblock {\em arXiv preprint arXiv:1907.05019}, 2019.

\bibitem{ba2016layer}
Jimmy~Lei Ba, Jamie~Ryan Kiros, and Geoffrey~E. Hinton.
\newblock Layer normalization.
\newblock 2016.

\bibitem{bachlechner2020rezero}
Thomas Bachlechner, Bodhisattwa~Prasad Majumder, Huanru~Henry Mao, Garrison~W
  Cottrell, and Julian McAuley.
\newblock Rezero is all you need: Fast convergence at large depth.
\newblock {\em arXiv preprint arXiv:2003.04887}, 2020.

\bibitem{bapna2018training}
Ankur Bapna, Mia~Xu Chen, Orhan Firat, Yuan Cao, and Yonghui Wu.
\newblock Training deeper neural machine translation models with transparent
  attention.
\newblock {\em arXiv preprint arXiv:1808.07561}, 2018.

\bibitem{bapna-firat-2019-simple}
Ankur Bapna and Orhan Firat.
\newblock Simple, scalable adaptation for neural machine translation.
\newblock In {\em Proceedings of the 2019 Conference on Empirical Methods in
  Natural Language Processing and the 9th International Joint Conference on
  Natural Language Processing (EMNLP-IJCNLP)}, pages 1538--1548, Hong Kong,
  China, November 2019. Association for Computational Linguistics.

\bibitem{conneau2019unsupervised}
Alexis Conneau, Kartikay Khandelwal, Naman Goyal, Vishrav Chaudhary, Guillaume
  Wenzek, Francisco Guzm{\'a}n, Edouard Grave, Myle Ott, Luke Zettlemoyer, and
  Veselin Stoyanov.
\newblock Unsupervised cross-lingual representation learning at scale.
\newblock {\em arXiv preprint arXiv:1911.02116}, 2019.

\bibitem{fan2019reducing}
Angela Fan, Edouard Grave, and Armand Joulin.
\newblock Reducing transformer depth on demand with structured dropout.
\newblock {\em arXiv preprint arXiv:1909.11556}, 2019.

\bibitem{grefenstette2019generalized}
Edward Grefenstette, Brandon Amos, Denis Yarats, Phu~Mon Htut, Artem Molchanov,
  Franziska Meier, Douwe Kiela, Kyunghyun Cho, and Soumith Chintala.
\newblock Generalized inner loop meta-learning.
\newblock {\em arXiv preprint arXiv:1910.01727}, 2019.

\bibitem{guo2019spottune}
Yunhui Guo, Honghui Shi, Abhishek Kumar, Kristen Grauman, Tajana Rosing, and
  Rogerio Feris.
\newblock Spottune: transfer learning through adaptive fine-tuning.
\newblock In {\em Proceedings of the IEEE Conference on Computer Vision and
  Pattern Recognition}, pages 4805--4814, 2019.

\bibitem{pmlr-v97-houlsby19a}
Neil Houlsby, Andrei Giurgiu, Stanislaw Jastrzebski, Bruna Morrone, Quentin
  De~Laroussilhe, Andrea Gesmundo, Mona Attariyan, and Sylvain Gelly.
\newblock Parameter-efficient transfer learning for {NLP}.
\newblock In {\em Proceedings of the 36th International Conference on Machine
  Learning}, volume~97 of {\em Proceedings of Machine Learning Research}, pages
  2790--2799, Long Beach, California, USA, 09--15 Jun 2019. PMLR.

\bibitem{huang2016deep}
Gao Huang, Yu~Sun, Zhuang Liu, Daniel Sedra, and Kilian~Q Weinberger.
\newblock Deep networks with stochastic depth.
\newblock In {\em European conference on computer vision}, pages 646--661.
  Springer, 2016.

\bibitem{jang2016categorical}
Eric Jang, Shixiang Gu, and Ben Poole.
\newblock Categorical reparameterization with gumbel-softmax.
\newblock {\em arXiv preprint arXiv:1611.01144}, 2016.

\bibitem{lepikhin2020gshard}
Dmitry Lepikhin, HyoukJoong Lee, Yuanzhong Xu, Dehao Chen, Orhan Firat, Yanping
  Huang, Maxim Krikun, Noam Shazeer, and Zhifeng Chen.
\newblock Gshard: Scaling giant models with conditional computation and
  automatic sharding.
\newblock {\em arXiv preprint arXiv:2006.16668}, 2020.

\bibitem{liu2019end}
Shikun Liu, Edward Johns, and Andrew~J Davison.
\newblock End-to-end multi-task learning with attention.
\newblock In {\em Proceedings of the IEEE Conference on Computer Vision and
  Pattern Recognition}, pages 1871--1880, 2019.

\bibitem{liu2020multilingual}
Yinhan Liu, Jiatao Gu, Naman Goyal, Xian Li, Sergey Edunov, Marjan
  Ghazvininejad, Mike Lewis, and Luke Zettlemoyer.
\newblock Multilingual denoising pre-training for neural machine translation.
\newblock {\em arXiv preprint arXiv:2001.08210}, 2020.

\bibitem{ott-etal-2019-fairseq}
Myle Ott, Sergey Edunov, Alexei Baevski, Angela Fan, Sam Gross, Nathan Ng,
  David Grangier, and Michael Auli.
\newblock fairseq: A fast, extensible toolkit for sequence modeling.
\newblock In {\em Proceedings of the 2019 Conference of the North {A}merican
  Chapter of the Association for Computational Linguistics (Demonstrations)},
  pages 48--53, Minneapolis, Minnesota, June 2019. Association for
  Computational Linguistics.

\bibitem{ott2018scaling}
Myle Ott, Sergey Edunov, David Grangier, and Michael Auli.
\newblock Scaling neural machine translation.
\newblock {\em arXiv preprint arXiv:1806.00187}, 2018.

\bibitem{pires2019multilingual}
Telmo Pires, Eva Schlinger, and Dan Garrette.
\newblock How multilingual is multilingual bert?
\newblock {\em arXiv preprint arXiv:1906.01502}, 2019.

\bibitem{poole2016exponential}
Ben Poole, Subhaneil Lahiri, Maithra Raghu, Jascha Sohl-Dickstein, and Surya
  Ganguli.
\newblock Exponential expressivity in deep neural networks through transient
  chaos.
\newblock In {\em Advances in neural information processing systems}, pages
  3360--3368, 2016.

\bibitem{post2018call}
Matt Post.
\newblock A call for clarity in reporting bleu scores.
\newblock {\em arXiv preprint arXiv:1804.08771}, 2018.

\bibitem{qi2018and}
Ye~Qi, Devendra~Singh Sachan, Matthieu Felix, Sarguna~Janani Padmanabhan, and
  Graham Neubig.
\newblock When and why are pre-trained word embeddings useful for neural
  machine translation?
\newblock {\em arXiv preprint arXiv:1804.06323}, 2018.

\bibitem{radford2019language}
Alec Radford, Jeffrey Wu, Rewon Child, David Luan, Dario Amodei, and Ilya
  Sutskever.
\newblock Language models are unsupervised multitask learners.
\newblock {\em OpenAI Blog}, 1(8):9, 2019.

\bibitem{rosenbaum2017routing}
Clemens Rosenbaum, Tim Klinger, and Matthew Riemer.
\newblock Routing networks: Adaptive selection of non-linear functions for
  multi-task learning.
\newblock {\em arXiv preprint arXiv:1711.01239}, 2017.

\bibitem{sennrich2015neural}
Rico Sennrich, Barry Haddow, and Alexandra Birch.
\newblock Neural machine translation of rare words with subword units.
\newblock {\em arXiv preprint arXiv:1508.07909}, 2015.

\bibitem{shaw2019meta}
Albert Shaw, Wei Wei, Weiyang Liu, Le~Song, and Bo~Dai.
\newblock Meta architecture search.
\newblock In {\em Advances in Neural Information Processing Systems}, pages
  11225--11235, 2019.

\bibitem{pmlr-v97-stickland19a}
Asa~Cooper Stickland and Iain Murray.
\newblock {BERT} and {PAL}s: Projected attention layers for efficient
  adaptation in multi-task learning.
\newblock In {\em Proceedings of the 36th International Conference on Machine
  Learning}, volume~97 of {\em Proceedings of Machine Learning Research}, pages
  5986--5995, Long Beach, California, USA, 09--15 Jun 2019. PMLR.

\bibitem{vandenhende2019branched}
Simon Vandenhende, Stamatios Georgoulis, Bert De~Brabandere, and Luc Van~Gool.
\newblock Branched multi-task networks: deciding what layers to share.
\newblock {\em arXiv preprint arXiv:1904.02920}, 2019.

\bibitem{vaswani2017attention}
Ashish Vaswani, Noam Shazeer, Niki Parmar, Jakob Uszkoreit, Llion Jones,
  Aidan~N Gomez, {\L}ukasz Kaiser, and Illia Polosukhin.
\newblock Attention is all you need.
\newblock In {\em Advances in neural information processing systems}, pages
  5998--6008, 2017.

\bibitem{wang2019learning}
Qiang Wang, Bei Li, Tong Xiao, Jingbo Zhu, Changliang Li, Derek~F Wong, and
  Lidia~S Chao.
\newblock Learning deep transformer models for machine translation.
\newblock In {\em Proceedings of the 57th Annual Meeting of the Association for
  Computational Linguistics}, pages 1810--1822, 2019.

\bibitem{wang2020balancing}
Xinyi Wang, Yulia Tsvetkov, and Graham Neubig.
\newblock Balancing training for multilingual neural machine translation.
\newblock {\em arXiv preprint arXiv:2004.06748}, 2020.

\bibitem{wei2020multiscale}
Xiangpeng Wei, Heng Yu, Yue Hu, Yue Zhang, Rongxiang Weng, and Weihua Luo.
\newblock Multiscale collaborative deep models for neural machine translation.
\newblock {\em arXiv preprint arXiv:2004.14021}, 2020.

\bibitem{wu2018blockdrop}
Zuxuan Wu, Tushar Nagarajan, Abhishek Kumar, Steven Rennie, Larry~S Davis,
  Kristen Grauman, and Rogerio Feris.
\newblock Blockdrop: Dynamic inference paths in residual networks.
\newblock In {\em Proceedings of the IEEE Conference on Computer Vision and
  Pattern Recognition}, pages 8817--8826, 2018.

\bibitem{yang2019xlnet}
Zhilin Yang, Zihang Dai, Yiming Yang, Jaime Carbonell, Russ~R Salakhutdinov,
  and Quoc~V Le.
\newblock Xlnet: Generalized autoregressive pretraining for language
  understanding.
\newblock In {\em Advances in neural information processing systems}, pages
  5754--5764, 2019.

\bibitem{zhang2019improving}
Biao Zhang, Ivan Titov, and Rico Sennrich.
\newblock Improving deep transformer with depth-scaled initialization and
  merged attention.
\newblock In {\em Proceedings of the 2019 Conference on Empirical Methods in
  Natural Language Processing and the 9th International Joint Conference on
  Natural Language Processing (EMNLP-IJCNLP)}, pages 897--908, 2019.

\bibitem{zhang2020improving}
Biao Zhang, Philip Williams, Ivan Titov, and Rico Sennrich.
\newblock Improving massively multilingual neural machine translation and
  zero-shot translation.
\newblock {\em arXiv preprint arXiv:2004.11867}, 2020.

\bibitem{zhang2019fixup}
Hongyi Zhang, Yann~N Dauphin, and Tengyu Ma.
\newblock Fixup initialization: Residual learning without normalization.
\newblock {\em arXiv preprint arXiv:1901.09321}, 2019.

\end{thebibliography}
\newpage
\appendix
\section{Gradient analysis}
We provide a detailed derivation of Eq. \ref{eq:grad}. The gradient backpropagated to layer $l$, $\frac{\partial \mathcal{L}}{\partial x_l} $, can be computed by applying the chain rule:
\begin{align}
\frac{\partial \mathcal{L}}{\partial x_l} &= \frac{\partial \mathcal{L}}{\partial x_L} \times \frac{ \partial x_L}{\partial x_l} \\
\end{align}
To compute $ \frac{ \partial x_L}{\partial x_l}$, we first apply Eq. \ref{eq:ll} recursively to expand $x_L$ as:
\begin{align}
    x_L &= x_{L-1} + z_{L-1}\times F_{L-1}(x_{L-1}) \\
        &= x_{L-2} + z_{L-2}\times F_{L-2}(x_{L-2}) + z_{L-1}\times F_{L-1}(x_{L-1}) \\
        &= x_l + \sum_{m=l}^{L-1} z_m F_m(x_m) \\
  \frac{ \partial x_L}{\partial x_l} &= 1 +      \sum_{m=l}^{L-1} z_m  \frac{ \partial F_m(x_m) }{\partial x_l}    \\
  \frac{\partial \mathcal{L}}{\partial x_l} &= \frac{\partial \mathcal{L}}{\partial x_L} \times (1 +      \sum_{m=l}^{L-1} z_m  \frac{ \partial F_m(x_m) }{\partial x_l}  )
\end{align}

\section{Training Details}

\subsection{Training procedure}
\label{app:training}
The proposed training procedure is motivated by the Generalized Inner Loop Meta-learning \cite{grefenstette2019generalized} although we use first-order gradient as approximation. Specifically, we treat $q_{\phi}$ as ``meta parameters" and the rest of the Transformer parameters $\Theta$ as ``task-specific" parameters. A key difference is that in our case there is only one task and the support set and target set are from the same distribution. At a high-level, we learn $\Theta$ in an inner-loop while updating $q_{\phi}$ from the unrolled gradient steps. Such nested optimization is computationally expensive as the graph for multiple steps needs to be stored in memory, and training was found to be unstable due to challenges in backpropagating second-order gradients through multiple steps \cite{antoniou2018train}. We adopt a multi-step loss approximation using first-order gradients only as is shown to be effective in \cite{antoniou2018train}. Specifically, in each outer loop we take the latest parameters of $q_{\phi}^{t-1}$, and perform $I$ inner loop steps. The gradients from each inner loop loss $\hat{\mathcal{L}}$ are directly backpropagated to $\Theta$, and the last step's gradient are used to update $q_{\phi}$, which is a special case of multi-step loss annealing where $\omega_{I-1}=1$, $\omega_j=0$ for $j<I-1$. 
\begin{algorithm}[H]
\caption{\label{alg:main} Training with latent layers in multilingual setting}
\begin{algorithmic}[1]
    \State{\textbf{Input:} training examples from $N$ languages $\{D_n\}_{n=1}^N$; total number of training steps $T$; inner loop update frequency $I$}
    \State{}{Initialize  $\Theta$, $q_{\phi}^0 = \{\alpha_l^0\}$; $t=0.$ }
    \For{t=1, ..., T}
        \For{i=1, ..., I}
            \For{n = 1, ..., N}
                \State{Sample a mini-batch $(x, y) \sim D_n$}.
                \State{Compute $z_{l=0, ..., L-1}$ all at once following Eq. \ref{eq:z_sample} with samples $\epsilon_l \sim \mathcal{G}$  }
                \State{Compute loss $\hat{\mathcal{L}}_{LL}((x, y);\Theta_{i-1}, q_{\phi}^{t-1})$ with Eq. \ref{eq:ll_loss}}
    
            \EndFor
             \State{Update $\Theta_i=\Theta_{i-1} - \eta \nabla_{\Theta_{i-1}} \hat{\mathcal{L}}_{LL}$ }
        \EndFor
		\State{Update $q_{\phi}^t=q_{\phi}^{t-1} - \eta \nabla_{q_{\phi}^{t-1}} \hat{\mathcal{L}}_{LL} $}
	\EndFor
\end{algorithmic}
\end{algorithm}
\subsection{Training stability.} 

\begin{wrapfigure}{r}{0.6\textwidth}
\centering
\includegraphics[width=\linewidth]{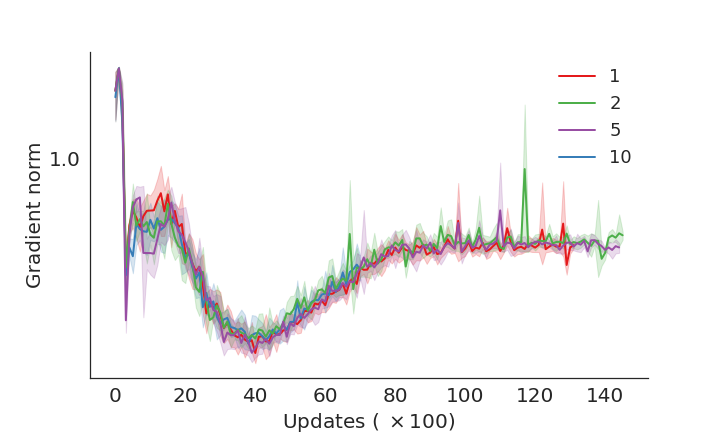}    
\caption{Comparison of gradient norms using different inner loop iterations $I$ to verify training stability is not sensitive to the choice of $I$.}
\label{fig:inner_loop}
\vspace{-20mm}
\end{wrapfigure}

We examine the stability of our training procedure, e.g. whether training is sensitive to the choice of inner loop frequencies. Figure \ref{fig:inner_loop} plots the gradient norms of using $I\in \{1,2,5,10\}$, and the impact on translation performance is summarized in Table \ref{tab:inner}.

\section{Experiments Implementation Details}

\subsection{Dataset description}
\label{app:data}
For WMT'16 English-German experiment, we used the same preprocessed data provided by \cite{wang2019learning} \footnote{The authors of \cite{wang2019learning} provided the downloadable data at \url{https://drive.google.com/uc?export=download&id=0B_bZck-ksdkpM25jRUN2X2UxMm8}}, including the same validation (\textit{neewsteest2013}) and test (\textit{neewsteest2014}) splits. The data volume for train, validation and test splits are 4500966, 3000, 3003 sentence pairs respectively.  The data was tokenized and numberized with a joint BPE (byte pair encoding) \cite{sennrich2015neural} vocabulary with 32k merge operations.  

For multilingual translation experiments, we use the same preprocessed data\footnote{The authors of \cite{wang2020balancing} provided the downloadable data at \url{https://drive.google.com/file/d/1xNlfgLK55SbNocQh7YpDcFUYymfVNEii/view?usp=sharing}} provided by \cite{wang2020balancing} using the same train, valid, and test split as in \cite{qi2018and}. The data volumes for related and diverse language groups are summarized in Table \ref{tab:app_data}.

For crosslingual language modelling we used data from Wikipedia from the 25 languages used in the mBART~\cite{liu2020multilingual} model, using the same data collection and preprocessing as \cite{conneau2019unsupervised}. We list the languages used and corresponding Wikipedia corpus size in Table~\ref{tab:datastats}. A random sample of 5000 sentences from each of the languages was used as held-out data to compare models. 
\begin{table*}[ht]
\small
    \centering
    \begin{tabular}{l|ccccccccc}
    \toprule
     &  Avg.  & bos & mar &  hin & mkd &  ell & bul &  fra &  kor  \\
    \hline
    $I=1$  & 28.28 &	23.0 & 14.1 & 19.2 & 31.6 & 37.2 & 39.4 & 40.8 & 20.9 \\
$I=2$  & 28.49 &	23.4 & 15.1 & 19 & 32.1 & 37.2 & 39.4 & 40.8 & 20.9 \\
$I=5$  & 28.24 &	23.4 & 14.5 & 18.6 & 32.1 & 37.2 & 39.1 & 40.2 & 20.8 \\
$I=10$  & 28.25 &	23.8 & 14.3 & 19 & 314 & 37 & 39.4 & 40.5 & 20.6 \\
    \bottomrule
    \end{tabular}
    \newline
    \caption{BLEU scores on validation set to assess the impact of the inner loop frequency $I$ on training stability and model performance, evaluated on the M2O-Diverse dataset.}\label{tab:inner}
\end{table*}

\begin{table}[t]
\begin{center}
\small
\begin{tabular}[b]{llr}
\toprule
\textbf{Code} & 
\textbf{Language} & 
\textbf{Sentences (M)}   \\
\midrule
{\bf En }& English & 41.9   \\
{\bf Ru }& Russian & 12.0 \\
{\bf Vi }& Vietnamese & 3.7\\
{\bf Ja }& Japanese & 1.7\\
{\bf De}& German & 16.7\\
{\bf Ro }& Romanian & 1.8\\
{\bf Fr }& French & 14.8 \\
{\bf Fi }& Finnish &  2.4\\
{\bf Ko }& Korean & 2.1\\
{\bf Es }& Spanish & 10.9\\
{\bf Zh } & Chinese (Sim) & 5.2 \\
{\bf It }& Italian & 9.7\\
{\bf Nl }& Dutch & 7.7\\
{\bf Ar }& Arabic &  3.2\\
{\bf Tr }& Turkish & 1.8\\
{\bf Hi }& Hindi & 0.6\\
{\bf Cs }& Czech &  2.7 \\
{\bf Lt }& Lithuanian & 0.9 \\
{\bf Lv }& Latvian & 0.45\\
{\bf Kk }& Kazakh & 1.0\\
{\bf Et }& Estonian & 2.2\\
{\bf Ne }& Nepali & 0.1\\
{\bf Si }& Sinhala & 0.1\\
{\bf Gu }& Gujarati & 0.1 \\
{\bf My }& Burmese & 0.4\\
\bottomrule
\end{tabular}
\caption{A list of the 25 languages and corresponding Wikipedia corpus size (in millions of sentences) used for crosslingual masked language modelling.}
\label{tab:datastats}
\end{center}
\end{table}

\subsection{Models and hyperparameters}
\label{app:impl}
Both baselines and proposed models are implemented using Transformer models in fairseq \cite{ott-etal-2019-fairseq}. For baseline models, we use the pre-norm setting  
 which provides a stronger baseline since it was shown to more effective for training deeper Transformer models than post-norm\cite{ott2018scaling,wang2019learning}. Therefore, the comparison with baseline can focus on evaluate the difference made from using latent layers. We use per-token negative loglikelihood (NLL) loss on the validation set to choose the loss coefficients for $\beta$ and $\lambda$.

\setlength{\tabcolsep}{2.5pt}
\begin{table*}[h]
\small
    \centering
    \begin{tabular}{l|cccccccc|cccccccc}
    \toprule
    & \multicolumn{8}{c|}{\bf Related} & \multicolumn{8}{c}{\bf Diverse} \\
      &  \bf aze & \bf bel & \bf glg  &  \bf slk  & \bf  ces &  \bf por &  \bf rus &  \bf  tur & \bf bos & \bf mar &  \bf hin & \bf mkd &  \bf ell & \bf bul &  \bf fra &  \bf kor \\
    \hline
 {\bf train (K)}  & 5.94 &	4.51  & 10  & 61.5 & 103 & 195 & 208 & 182 & 5.64 &	9.84 & 18.79 & 25.33 & 134 & 174 & 192 & 205  \\
{\bf valid} & 671 &	248 & 682 & 2271& 3462  & 4035 & 4814  & 4045 & 474 &	767 & 854 & 640 & 3344 & 4082 & 4320 & 4441  \\
{\bf test}  & 903 &	664  & 1007 & 2445 & 3831 & 4855 & 5483  & 5029 & 463 &	1090 & 1243 & 438 & 4433 & 5060 & 4866 & 5637  \\
    \bottomrule
    \end{tabular}
    \newline
    \caption{Data statistics (number of sentence pairs or thousands of sentence pairs for training data) for languages used in multilingual translation experiments.}\label{tab:app_data}
\end{table*}

\paragraph{WMT'16 English-German.} All models were trained for 75 epochs and evaluating on the last checkpoint. For Transformer-big, we use the standard model architecture as is described in \cite{vaswani2017attention}: $d=1024$ for embedding and hidden dimension, and $d=4096$ for FFN dimension, 6-layer encoder and decoder, 0.3 dropout (0.1 after attention sub-layer and ReLU activation). Model was trained with 8192 token per GPU and 32 GPUs, learning rate 7e-4 and 8000 warm-up updates with Adam optimizer. For deeper models, i.e. both DLCL (baseline) and latent layers (LL, the proposed approach), since the depth is increased we reduce the model width by using  $d=512$ for embedding and hidden dimension, and $d=1024$ for FFN dimension, and 4 attention heads. Also, we found for deeper models we were able to use almost $2\times$ learning rate (1.5e-3). We use $\beta=1$ and $\lambda=0.1$ for latent layers models.

\paragraph{Crosslingual Masked Language Modelling.} We use the XLM-R$_{\mathrm{Base}}$ architecture of \cite{conneau2019unsupervised}, which has a hidden dimension of 768, but we explore increasing the number of layers, considering 24, 48 and 96 layer models. We learn a Sentencepiece vocabulary of size 40k on the training data.  We evaluate the models after 100k updates (as opposed to \cite{conneau2019unsupervised} who train for 1.5 million updates) with a per-GPU batch size of 8192 tokens and 32 GPUs. Note we do not use language-aware latent variables despite the multilingual training data. We use the Adam optimizer with learning rate of either 5e-4, 2.5e-4 (24 or 48 layers) or 1.25e-4 (96 layers) and linear warmup followed by polynomial decay with either 5000 (24 or 48 layers) or 15000 (96 layers) warmup steps. For our static model with 96 layers we further tried increasing warmup to 30000 steps and decreasing the learning rate to 1.5625e-5 but this did not help with training loss divergence issues. When using LayerDrop we use 50\% dropout probability. We re-use all other hyperparameters from XLM-R \cite{conneau2019unsupervised} (i.e.\ token masking probability etc.).

 \paragraph{Multilingual Machine Translation.} For multilingual experiments, we use a single Transformer network shared across all languages for both the encoder and decoder, with the same model size as used in \cite{wang2020balancing}. We use the same optimization hyperparameters (learning rate, warm up schedule, etc) as used in WMT English-German experiments except that the batch size is  4096 tokens  per-language and we train the model for 14k updates, and evaluated on the last checkpoint. Similarly, we use beam search with beam size 5 and length penalty 1.0 for decoding. 

\section{Visualizations}
\label{app:vis}
We provide visualizations of the layer selection samples $z_l$ to further illustrate modeling choices  around $\mathcal{L}_K$ and priors. 

\begin{figure}[h]
\vspace{-2mm}
\begin{center}
\includegraphics[width=\textwidth]{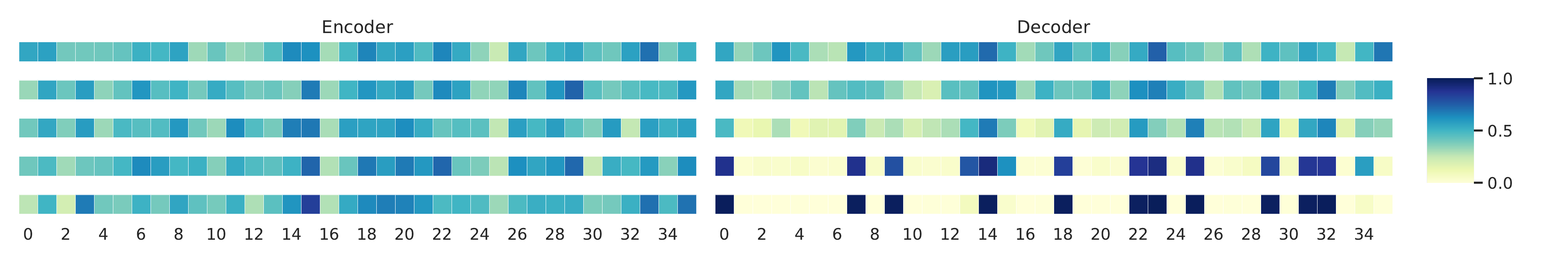}
\caption{Layer selection samples throughout training evaluated on the WMT'16 English-German dataset. Rows correspond to samples from encoder and decoder with 36 latent layers at epoch 2, 6, 25, 50, and 100 respectively. $\mathcal{L}_K$ ($K=12$) was applied to decoder only and not encoder to contrast the discretizing and pruning effect. }
\label{fig:en-de-dec-k}
\end{center}
\vskip -0.2in
\end{figure}

\paragraph{Effect of $\mathcal{L}_K$. } First, we show that adding the auxiliary loss $\mathcal{L}_K$ discretizes the samples and achieve the pruning purpose by enforcing sparsity of the resulting model.  In Figure \ref{fig:en-de-dec-k}, we visualized samples throughout training using the WMT'16 English-German dataset. Since decoder depth directly contributes to latency at inference time, we only apply $\mathcal{L}_K$ with $K=12$ to latent layers training in decoder and not in encoder. We could see that samples $z_l$ in decoder becomes discrete throughout training while samples in encoder stay continuous.

\begin{figure}[h]
\vspace{-2mm}
\begin{center}
\centerline{\includegraphics[width=\textwidth]{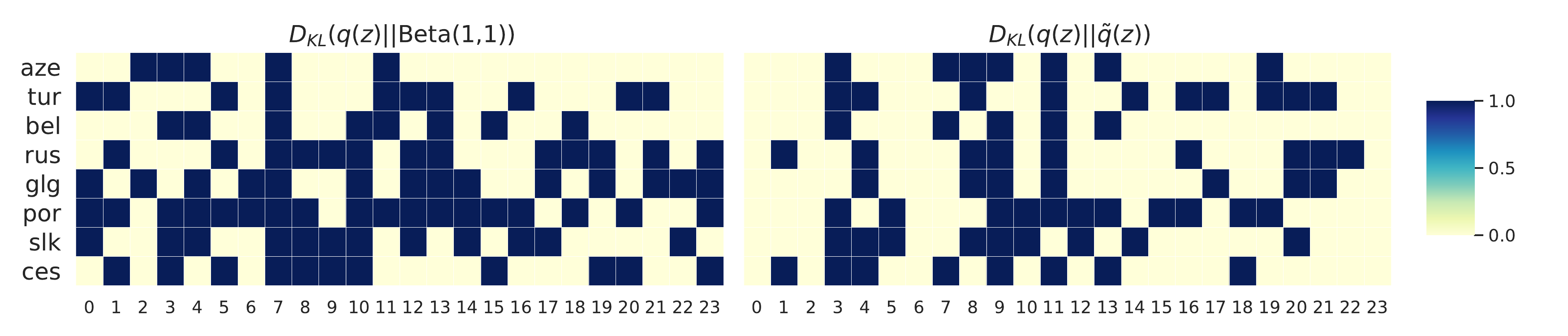}}
\caption{Layer selection samples $z_l$ at inference time trained with uniform prior (left) and aggregated posterior $\tilde{q}(z)$ (right) in $\KL$. Compared to the uniform prior, using aggregated posterior is more effective for ``pruning" by encouraging consistent ``select" and ``skip" across languages. For example, layer 0, 2, 6, and 23 can be complete pruned for all languages besides language-specific pruning (e.g. for each language/row, layers corresponding to lighter cells could be pruned to derive a sub-network for the given language). This property is appealing for deploying one shared multilingual model for all languages.}
\label{fig:samples_prior}
\end{center}
\vskip -0.2in
\end{figure}

\paragraph{Effect of priors.} In Section \ref{sec:analysis} we showed the difference between using an uniform prior Beta(1,1) and aggregated posterior $\tilde{q}(z)$ in the early stage of training. In Figure \ref{fig:samples_prior}, we further compared the resulting samples used at inference time, where we can see that using aggregated posterior $\tilde{q}(z)$ leads to more consistent sampling behavior for each layer (either ``all select" or ``all skip") across languages and thus obtain increased sparsity and a more compact model. We used the O2M-Related language group for evaluation, where we could observe qualitatively common layer selection patterns for languages of the same language family, e.g. aze (Azerbaijani) and tur (Turkish), bel (Belorussian) and rus (Russian), glg (Galician) and por (Portuguese), slk  (Slovak) and ces (Czech). We leave a systematic study of layer selection and linguistic similarity to future work.


\end{document}